# Brain Tumor Detection and Classification based on Hybrid Ensemble Classifier

**Ginni Garg[1], Ritu Garg**

*Department of Computer Engineering*

*National Institute of Technology, Kurukshetra, 136119*

*gargginni01@gmail.com, ritu.59@gmail.com*

**Abstract:** *To improve patient survival and treatment outcomes, early diagnosis of brain tumors is an essential task. It is a difficult task to evaluate the magnetic resonance imaging (MRI) images manually. Thus, there is a need for digital methods for tumor diagnosis with better accuracy. However, it is still a very challenging task in assessing their shape, volume, boundaries, tumor detection, size, segmentation, and classification. In this proposed work, we propose a hybrid ensemble method using Random Forest (RF), K-Nearest Neighbour, and Decision Tree (DT) (KNN-RF-DT) based on Majority Voting Method. It aims to calculate the area of the tumor region and classify brain tumors as benign and malignant. In the beginning, segmentation is done by using Otsu's Threshold method. Feature Extraction is done by using Stationary Wavelet Transform (SWT), Principle Component Analysis (PCA), and Gray Level Co-occurrence Matrix (GLCM), which gives thirteen features for classification. The classification is done by hybrid ensemble classifier (KNN-RF-DT) based on the Majority Voting method. Overall it aimed at improving the performance by traditional classifiers instead of going to deep learning. Traditional classifiers have an advantage over deep learning algorithms because they require small datasets for training and have low computational time complexity, low cost to the users, and can be easily adopted by less skilled people. Overall, our proposed method is tested upon dataset of 2556 images, which are used in 85:15 for training and testing respectively and gives good accuracy of 97.305%.*

**Keyword:** Otsu's Threshold; SWT; PCA; GLCM; hybrid ensemble Classifier (KNN-RF-DT) based on Majority Voting method.

## 1. Introduction

A brain tumor is a cancerous or non-cancerous growth of abnormal cells in the brain, which leads to benign or malignant brain tumors. Most of the researchers are engaging in the primary type of tumor such as Gliomas. We have some ways to treat gliomas such as chemotherapy, radiotherapy, and surgery. Automation by computer-aided devices can be used to obtain the necessary clinical data such as tumor presence, location, and type. However, it is still a very challenging task in assessing their shape, volume, boundaries, tumor detection, size, segmentation, and classification. Also, brain tumor intensity varies from individual to individual. Magnetic Resonance Imaging (MRI) is preferred over other treatment and diagnosis methods because it gives superior image contrast in soft tissues and has non-invasive property. On applying different pulse sequences, we obtain a different type of MRI scans, such as (1) T1 weighted scans that distinguish between tumor and healthy tissues. (2) T2 weighted scans cause delineation of the edema region, and ultimately we get a bright image region. (3) T4-Gd scans which gives bright signal at tumor border by using a contrast agent. (4) FLAIR scans differentiate between cerebrospinal fluid (CSF) and edema region by using a signal of water molecule suppression. It is a difficult task to do annotation of brain tumors from MRI scans manually. Hence, there is a strong need for automation of brain tumor segmentation and classification with the help of computer vision and machine learning algorithms. Today, researchers are working on computer vision and machine learning algorithms for brain tumor segmentation and classification. Clinician's plans are highly expensive because they depend on various imaging techniques such as PET, MRI, and CT. The clinical methods provide extract pertinent information and comprehensive analysis from images. Computational techniques help to investigate the details present in medical images. Imaging methods can be used to find the position of brain tumors. MRI provides more meaningful information in contrast to other imaging modalities like CT.

The challenging task in Brain Tumor is due to high variability and inherent MRI data characteristics, e.g., variability in tumor sizes or shapes, tumor detection, area calculation, segmentation, classification, and finding uncertainty in segmented region. The most significant task in image understanding is image segmentation because it helps in feature extraction, area calculation, and significance in many real-life applications. It can be used, for example, estimation of tumor volume, tissue classification, blood cell delineation, and localization of tumors, matching of an atlas, surgical planning, and image registration. For monitoring oncologic therapy, the accurate and morphology quantification of tumors is a critical task. However, extensive scale work has been performed in this field; but still; clinicians depend on manual determination of tumor, due to lack of link between researchers and clinicians.

Recently, many techniques have been proposed for automatic brain tumor classification that can be categorized into machine learning (ML) and deep learning (DL) techniques based on the feature selection and learning mechanism. In ML approaches, feature selection and extraction is essential for classification. However, DL approaches extract and learn the features from the image directly. Recent DL approaches, particularly CNN provides good accuracy and is widely used in medical image analysis. Moreover, they have disadvantage over traditional methods (ML) as they need large dataset for training, have high time complexity, less accurate for applications where we have availability of small dataset and require expensive GPUs which ultimately increases cost to the users. Additionally, selecting the right deep learning tools is also a challenging task as it needs knowledge regarding various parameters, training method, and topology. On the other hand, machine-learning approaches have played key role in the area of medical imaging. Several learning based classifiers have already been used for classification and detection of brain tumors, which includes – support vector machine (SVM), artificial neural network (ANN), sequential minimal optimization (SMO), fuzzy C mean (FCM), Naïve Bayes (NB), Random Forest (RF), Decision Tree (DT) and K-Nearest Neighbor (KNN). KNN implementation is very simple and takes less computation and space complexity. It requires very less parameters to tune. The biggest advantage of DT is that it goes through all the outcomes of decision and finally traces each path to reach the conclusion. It is versatile; no complex mathematics involved, which makes it easy to understand. Further, Random Forest is itself ensemble classifiers of DT. It runs effectively on large dataset, which provides good parameter values for accuracies, precision and other evaluation metrics. Overall, these classifiers have received considerable research attention, as they require small dataset for training, low computational time complexity, low cost to the users, and can be easily adopted by less skilled people. Thus, in the present study, we work on hybrid ensemble classifiers in order to improve the accuracy of results obtained. Further, comparative study of various classifiers such as SVM, KNN, DT, RF, NB, ANN and proposed hybrid ensemble classifier is done.

The outline of the paper is as follows: Section 2 describes the related work; Section 3 describes the proposed method for area calculation and brain tumor classification. Section 4 gives information about experimental implementation and results, and Finally, Section 5 represents conclusion.

## 2. Related Work

Brain Tumor research has been conducted in various private multinational companies like, Siemens, Becton Dickinson, Medtronic, Accenture, GE Medical Systems, Atlantic Biomedical P. Ltd, and others. Both theoretical and experimental works of International arena are reported in the literature. Some of work done by good researchers is described below:

Othman [4] et al. proposed a method in which feature extraction is done by using Daubechies wavelets with DWT from MRI images. Each image consists of 17,689 feature vectors. Finally, classification is done using RBF a SVM kernel function. Sindhumol et al. [4] presents a spectral clustering (SC) technique for classification of brain tumor. Images (MRI) are divided into different clusters using the spectral distance. Feature reduction is made by using ICA and classification by using SVM. The training and testing data consists of 40 normal and 20 abnormal MRI images. Abd-Ellah [6] et al. proposed method consist of preprocessing of MRI images is done with the help of Median filters. DW does feature extraction, and PCA is used for feature reduction. Finally, classification is done by the SVM classifier using the RBF kernel function. The database consists of 80 images. SVM is trained using 43 abnormal and 5 normal images, and testing is done by using 27 abnormal images and 5 normal images. H. Kalbkhani [7] et al., proposed the subband of the detail coefficients and 2D DWT using (GARCH), Feature reduction is made from 61440 to 24 features. Feature extraction is done by linear discriminate analysis (LDA), which is further reduced using PCA. Finally, detection is done by using SVM and KNN identifier. The training and testing data consists of MRI images, normal and abnormal in ratio 10 to 70. The testing set consists of 7 normal and 49 abnormal images, while the training set contains 3 normal and 21 abnormal images. Saritha et al. [8] proposed classification technique for normal or abnormal brain tumor images considering 23 images for testing and 50 for training. Deep and Devi [9] et al., proposed a system in which the statistical method is used for texture feature extraction, neural network, and BPNN is used in segmentation and uncovering stages. The database consists of 42 images, which are further divided into training and testing as 30 and 12, respectively. Chandra [10] et al. proposed a new clustering algorithm based on PSO optimization with the help of MRI images. The clusters and corresponding centroids are being found out by algorithm, among them global best is considered. The dataset consists of 62 normal MRI images and 110 abnormal ones. Xuan and Liao [11], proposed a tumor detection method considering features of 3-types texture-based, intensity-based and symmetry-based. Then, total 40 features consisting of 13 intensity-based, 26 texture-based and 1 symmetry-based features are selected. Feature extraction is done from different images with 12 features from T2 images, 9 from T1 images and 19 from FLAIR images. The dataset contains 10 patients with 3 volumes each with 24 slices of MRI images. They divided the dataset equally into testing and training sets. Dhanalakshmi [12] et al., proposed work consist of k-means clustering for segmentation and then area is calculated using formula sqrt(P)*.264, where P is the no. of pixel with value 1. The proposed algorithm shows the reproducibility and good performance. Kaushik [13] et al., proposed method consists of segmentation using genetic algorithm. The corners of the brain tumor region are also extracted based on proposed algorithm. Rani [16] et al., proposed a method for MRI brain tumor image classification using SVM and segmentation using otsu's thresholding method. This paper compared its proposed work with KIFCM, K-means an Fuzzy c-means but their accuracy and executive time was more effective then all remaining existing methods.

Additionally, many deep learning models have been investigated recently for brain tumor detection and classification and achieved the competitive results. Chaudhary [15] et al., proposed a method based on deep learning for the segmentation of MRI brain tumor images. In which pre processing of MRI images was done using intensity normalization. Kamboj [17] et al., reviewed deep learning methods, which has advantages over traditional methods. Their focus is on design of architecture as compare to segmentation and feature extraction. Deep learning methods provide good accuracy but they required more computation time, space and dataset as compare to the traditional classifiers. However, traditional machine learning methods are easy to understand, interpret and required less space, dataset and computational cost in terms of hardware.

Moreover, none of the above mentioned machine learning approaches work on feature extraction using 3-fold techniques such as SWT+PCA+GLCM, which significantly increases the robustness of extracted features as SWT helped in capturing the abrupt changes of images. PCA reduced the dimensionality of input images from SWT, which reduced space and time complexity up to some extent. Finally, GLCM extracted various useful features from dimensionally reduced images of PCA. In addition, none of the above methods worked on hybrid ensemble classifiers, which helped in achieving good evaluation metrics using traditional classifiers, as best properties of each classifier add up to gave excellent results. Therefore, overall 3-fold robust feature extraction and hybrid ensemble classification is the main focus of interest in this present study, which helped in improving the various evaluation metrics and reduced space and time complexity using traditional classifiers.

## 3. Proposed Method

The proposed work aims at improving the performance of traditional classifiers. These classifiers require small datasets for training and have low computational time complexity thus appropriate for computer assisted brain tumor diagnosis and classification. We propose a hybrid ensemble method using KNN, Random Forest (RF) and Decision Tree (DT) (KNN-RF-DT) based on Majority Voting Method. It aims to calculate the area of the tumor region and classify brain tumors as benign and malignant. In the beginning, MRI images segments using the Otsu's Threshold method. Feature Extraction is done by Stationary Wavelet Transform (SWT), Principle Component Analysis (PCA) and Gray Level Co-occurrence Matrix (GLCM), which gives thirteen features for classification. The classification is done by hybrid ensemble classifiers (KNN-RF-DT) based on the Majority Voting method.

In the current research work, we have done comparative studies of various classifiers such as SVM, KNN, DT, RF, NB, ANN and proposed hybrid ensemble Classifier. Overall, it aimed at improving the performance by using traditional classifiers. Working of proposed implementation is shown in Fig.1.

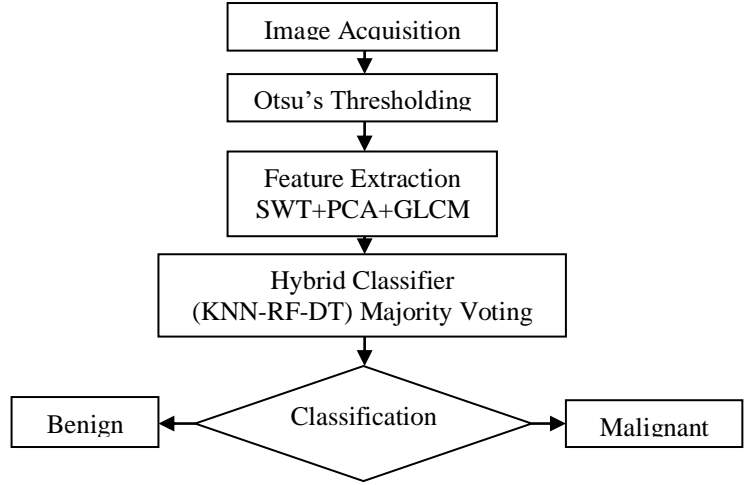

**Fig. 1** Flow diagram of proposed work

### a. Otsu's Method

Otsu method [19] is used for the automatic image threshold into two classed, foreground, and background based on a single threshold value. This threshold determines by maximizing inter-class variance and minimizing intra-class intensity variance. Threshold that minimizes the intra-class variance that describes by corresponding sum of variance of two classes:

$$\alpha_w^2(\mathbf{n}) = A_0(\mathbf{n}) * \alpha_0^2(\mathbf{n}) + A_1(\mathbf{n}) * \alpha_1^2(\mathbf{n}), \tag{1}$$

where $A_0, A_1$ – Probability of two classes, **n**=154- Threshold, $\alpha_0^2, \alpha_1^2$ - Variance of two classes. The Class Probability can be computed from the number of bins (L=256) as shown below:

$$A_0(\mathbf{n}) = \sum_{k=0}^{n-1} p(\mathbf{k}) \tag{2}$$

$$A_1(\mathbf{n}) = \sum_{k=n}^{L-1} p(\mathbf{k}), \tag{3}$$

where we know that minimum intra-class variance is equivalent to the maximum inter-class variance.

$$\alpha_b^2(\mathbf{n}) = \alpha^2 - \alpha_w^2(\mathbf{n}) = A_0 * (\beta_0 - \beta_T)^2 + A_1 * (\beta_1 - \beta_T)^2 = A_0(\mathbf{n}) * A_1(\mathbf{n}) * (\beta_0 - \beta_1)^2, \tag{4}$$

where $\beta_1(\mathbf{n})$, $\beta_0(\mathbf{n})$ and $\beta_T$ are class means

$$\beta_0(\mathbf{n}) = = \sum_{k=0}^{n-1} \mathbf{k} * p(\mathbf{k}) / A_0(\mathbf{n}) \tag{5}$$

$$\beta_1(\mathbf{n}) = = \sum_{k=n}^{L-1} \mathbf{k} * p(\mathbf{k}) / A_1(\mathbf{n}) \tag{6}$$

$$\beta_T = = \sum_{k=0}^{L-1} \mathbf{k} * p(\mathbf{k}) \tag{7}$$

$$A_0 * \beta_0 + A_1 * \beta_1 = \beta_T \tag{8}$$

$$A_0 + A_1 = 1 \tag{9}$$

Above computations, give us effective algorithm as class probability and class means computed iteratively. Histogram of brain tumor image with bins 256 and threshold 154 is shown below in Fig.2.

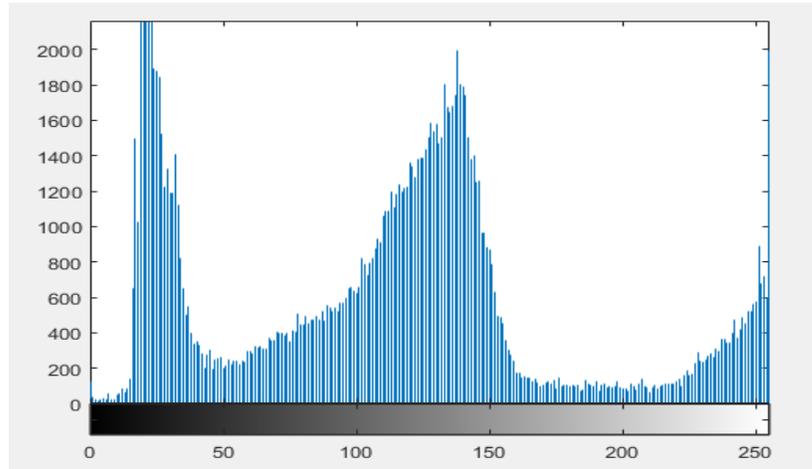

**Fig. 2** Histogram of brain tumor image with bins L= 256 and threshold n= 154

### b. Stationary Wavelet Transform

As far as signal is concerned, slow changes can be captured with the help of Fourier Transform, whereas when the images undergo abrupt changes that can be captured with the concept of wavelets. Wavelet is a small oscillation whose frequency inversely varies with scaling. To capture abrupt changes, we need high frequency and small scaling so we need the concept of wavelet. Stationary Wavelet Transform (SWT) [20] algorithm is designed to overcome the problem of translation-invariance of the discrete wavelet transform (DWT). Translation-invariance can be achieved by removing the down-samples and up-samples in the DWT and up sampling the filter coefficients by a factor of $2^{(j-1)}$ in the $j^{th}$ level of algorithm. SWT has various applications such as - Signal de-noising, Pattern Recognition, Brain image classification, Pathological brain detection. In our proposed work, we work on 1-D SWT, where $j = 1$.

The origin of the wavelets and its types from Fourier transform is shown in Fig.3.

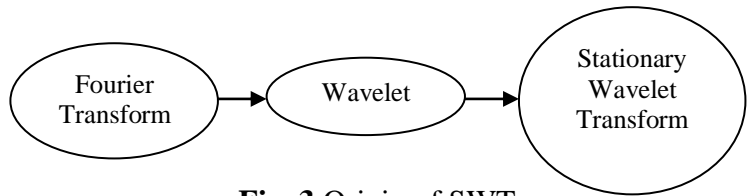

**Fig. 3** Origin of SWT

The digital implementation of SWT is shown below in Fig. 4 in which each level is an up-sampled version of the previous level.

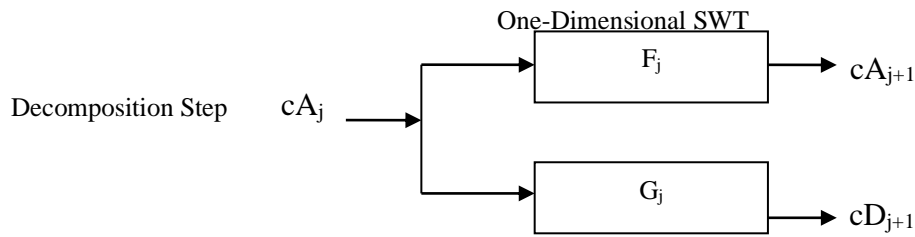

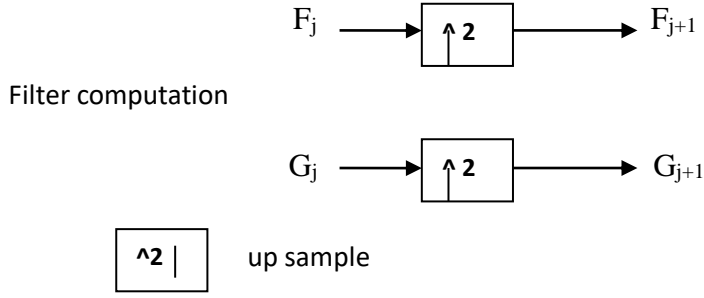

Initialization: $cA_0 = s$ and $F_0 = Lo\_D$ and $G_0 = Hi\_D$

**Fig. 4** Implementation of SWT in which each level is an up-sampled version of the previous level

### c. Principal Component Analysis

PCA [21] is an orthogonal transformation of correlated variables to linearly uncorrelated variables known as principal components. The first principal component has the highest variances; successive components contain the most significant possible variance such that it remains orthogonal to previous components. It is used to reduce the dimensions or features with which we have to train our classifier, which ultimately helps in reducing the time and space complexity for given data computation.

**First Component**

$$Y_{(1)} = \arg\max_{|Y|=1}\{\Sigma_k(t_1)_{(k)}^2\} = \arg\max_{|Y|=1}\{\Sigma_k(\mathbf{X}_{(k)}.Y)^2\}, \tag{10}$$

Where $Y_{(1)}$ denotes the unit vector and $\mathbf{X}$ is Image matrix.
Equation in matrix form:

$$Y_{(1)} = \arg\max_{|Y|=1}\{\|\mathbf{X}Y\|^2\} = \arg\max_{|Y|=1}\{Y^T\mathbf{X}^T\mathbf{X}Y\} \tag{11}$$

$$Y_{(1)} = \arg\max\{Y^T\mathbf{X}^T\mathbf{X}Y/Y^TY\} \tag{12}$$

Largest Eigen value of the matrix is Rayleigh Quotient, where $\mathbf{X}^T\mathbf{X}$ is positive semi-definite matrix and $Y$ = corresponding Eigen vector

**Further Components**

The Nth component can be found by subtracting the first **N**-1 principle component from **X**, where **N** = 13 for proposed method.

$$\mathbf{X}_N = X - \sum_{n=1}^{N-1}\mathbf{X}Y(n)Y(n)^T \tag{13}$$

Further, weight vector can be found as describe below:

$$Y_{(K)} = \arg\max_{\|Y\|=1}\{\|\mathbf{X}_K*Y\|^2\} = \arg\max\{Y^T\mathbf{X}_K^T\mathbf{X}_KY/Y^TY\} \tag{14}$$

### d. Gray-Level Co-occurrence Matrix

It is a statistical method that describes the spatial relationship of pixels based on the spatial grey-level dependence matrix. GLCM [22] calculates the texture of the image by calculating the frequency of corresponding pixels and their spatial relationships. All the thirteen features which are used in proposed work as follow contrast, correlation, energy, homogeneity, mean, standard deviation, kurtosis, skewness, variance, smoothness, IDM, RMS, entropy. There detailed information is given below:

$$\text{Contrast (}\mathbf{C}\text{)} = \sum_{t,r=1}^{T,R} |t-r|^2 \mathbf{q}(t,r), \tag{15}$$

where, $\mathbf{q}(t,r)$ is GLCM, t & r are row & column, T is total rows and R is total columns.

$$\text{Correlation (Corr)} = \sum_{t,r=1}^{T,R} \big((t-\mu)(r-\mu)\mathbf{q}(t,r)\big)/(\sigma(t)*\sigma(r)), \tag{16}$$

where, mean is μ and standard deviation is σ

$$\text{Energy (}\mathbf{E}\text{)} = \sum_{t,r=1}^{T,R} \mathbf{q}(t,r)^2 \tag{17}$$

$$\text{Homogeneity (}\mathbf{H}\text{)} = \sum_{t,r=1}^{T,R} \mathbf{q}(t,r)/(1+|t-r|) \tag{18}$$

$$\text{Mean (}\mu\text{)} = \frac{1}{T*R} * \sum_{t=1}^{T}\sum_{r=1}^{R} \mathbf{q}(t,r) \tag{19}$$

$$\text{Standard Deviation (}\sigma\text{)} = \sqrt[2]{\frac{1}{T*R} * \sum_{t=1}^{T}\sum_{r=1}^{R}(\mathbf{q}(t,r)-\mu)} \tag{20}$$

$$\text{Kurtosis (}\mathbf{K}\text{)} = \{\frac{1}{T*R} * \sum_{t=1}^{T}\sum_{r=1}^{R}((\mathbf{q}(t,r)-\mu)/\sigma)\wedge 4\} - 3 \tag{21}$$

$$\text{Skewness (}\sigma\text{)} = \sqrt[2]{\frac{1}{T*R} * \sum_{t=1}^{T}\sum_{r=1}^{R}(\mathbf{q}(t,r)-\mu)} \tag{22}$$

$$\text{Variance (}\mathbf{Var}\text{)} = \frac{1}{T*R} * \sum_{t=1}^{T}\sum_{r=1}^{R}(\mathbf{q}(t,r)-\mu) \tag{23}$$

$$\text{Smoothness (}\mathbf{R}\text{)} = 1-1/(1+\sigma^2) \tag{24}$$

$$\text{IDM (}\mathbf{HH}\text{)} = \sum_{t,r=1}^{T,R} \frac{\mathbf{q}(t,r)}{1+|t-r|} \tag{25}$$

$$\text{RMS (}\mathbf{y}\text{)} = \sqrt[2]{\sum_{t,r=1}^{T,R}(|\mathbf{q}(t,r)|)^2/T} \tag{26}$$

$$\text{Entropy (}\mathbf{h}\text{)} = -\sum_{t,r=1}^{T,R} \mathbf{q}(t,r)(\log \mathbf{q}(t,r)) \tag{27}$$

**e. Random Forest**

Random Forest [23] is an ensemble classifier formed by the fusion of many decision trees. It calculates the result on the basis of the majority voting method. Random forest is more superior then the decision tree as it overcomes the problem of over-fitting. As a tree grows deep, they start to over-fit, i.e., they have low bias and high variance. Random forest uses the different parts of the same training dataset on different trees and helps them averaging multiple decision trees and avoid over-fitting, which increases bias and reduces variance, which boosts performance. Internal working of random forest is shown in Fig.5. We are using 100 decision trees, which are trained using training data consisting of 2172 images by concept of bagging.

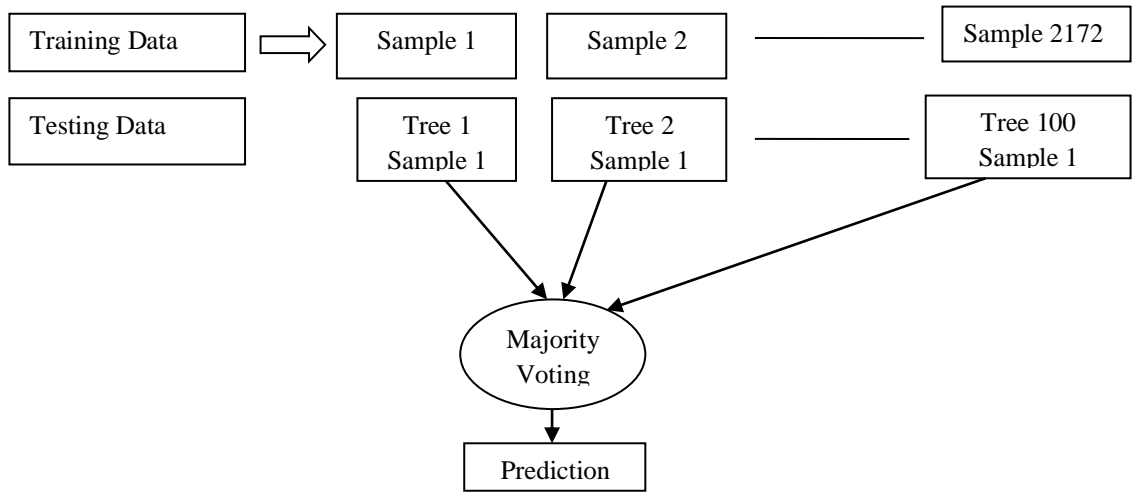

**Fig. 5** Internal working of Random Forest

### f. Decision Tree

Decision Tree [24] is like a conditional control statements, which performs the research operations such as decision analysis. There occurs the problem of over-fitting when trees become deep enough. It is like a tree structure, where each node represents attribute or feature on bases of which one can get the outcome. Each leaf node holds the information related to the class label. Working of decision tree is shown in Fig.6. Features are used as internal nodes of the tree and class are leaf nodes.

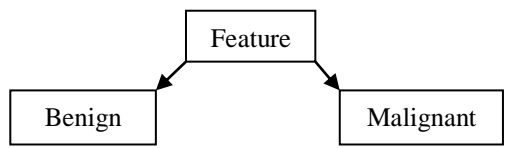

**Fig. 6** Working of Decision Tree

### g. K-Nearest Neighbor

KNN [25] is a lazy learning technique in which functions are locally approximation. It can be used for both classification and regression. In which weights assign to neighbors based on distance, i.e., if the distance is d, then weights assign as 1/d such that nearest neighbors contribute more then distant. It has little time complexity because training consists of just calculating the Euclidean distance. Euclidean Distance Measurement is shown in Fig.7. We have two classes, one represented by using square and other by triangle. Now, prediction for testing data circle is done based on minimum Euclidean distance measurement.

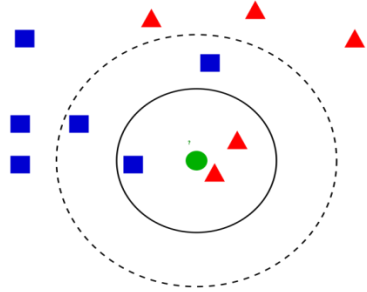

**Fig. 7** Euclidean Distance Measurement

### h. Hybrid Ensemble Classifier

Proposed hybrid ensemble classifier KNN-RF-DT shown in Fig. 8, where prediction is considered for at least two-ratio-one voting of classifiers to a specific class such as benign or malignant.

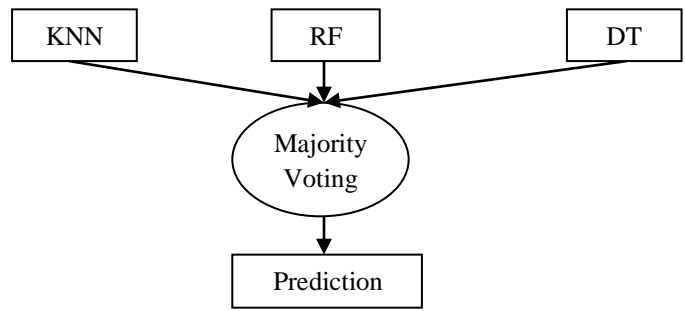

**Fig. 8** Proposed Hybrid Ensemble Classifier

---

**Alogrithm-1** Classification process with Hybrid Ensemble Classifier

1. Mdl1 ← Model of KNN with K=1
2. Mdl2 ← Model of Random Forest with 100 Trees
3. Mdl3 ← Model of Decision Tree
4. p1 ← predict from Mdl1
5. p2 ← predict from Mdl2
6. p3 ← predict from Mdl3
7. var right ← 0 and var left ← 0
8. if p1 is "Malignant" then
9.    right ← right+1
10. else
11.    left ← left+1
12. end

13. if p2 is "Malignant" then
14.     right ← right+1
15. else
16.     left ← left+1
17. end
18. if p3 is "Malignant" then
19.     right ← right+1
20. else
21.     left ← left+1
22. end
23. if right is greater then left then
24.     species ← "Malignant"
25. else
26.     species ← "Benign"
27. end

## 4. Implementation and Results

### a. Dataset used and various SWT filter's Matrix Representations

In the proposed work, Cancer Genome Atlas Glioblastoma Multi-forme (TCGA-GBM) [14] data collection is used to conduct the experimental computation of the proposed approach. This is an open and standard Glioblastma Multi-forme dataset, which is main type of brain tumor. It is available freely for research work and highly accurate dataset. Hence, no decision from any committee is required on this dataset. The augmentation process is also used to increase dataset such that we have an average of 2556 samples of T1-weighted images used to test the proposed approach, based on testing and training images in ratio 85:15. A distribution of dataset between testing and training is shown in Table.1 and image segmentation using Otsu's method and 1-D SWT filters is shown in Fig. 9.

**Table 1.** Database for Benign and Malignant classification

| Database | Training Dataset | Testing Dataset |
|---|---|---|
| Benign | 1086 | 192 |
| Malignant | 1086 | 192 |

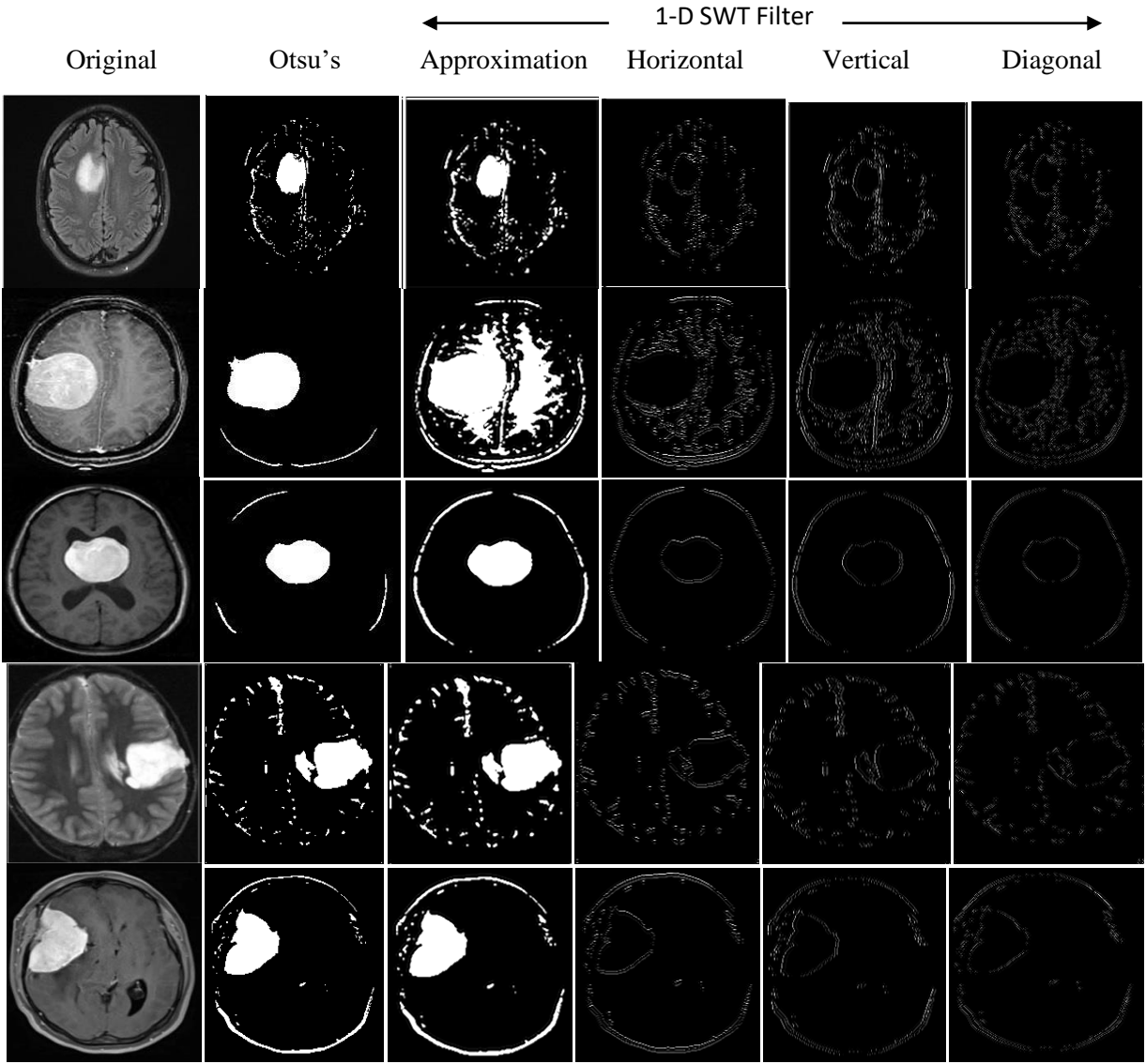

**Fig. 9** Results of Otsu's, 1-D SWT filters with Approximation, Horizontal, Vertical and Diagonal matrix.

b. Evaluation Metrics and its Graphical Representation

The following parameters such as Accuracy, Precision, Sensitivity, specificity, F1-score, and Youden index are calculated based on proposed methodology with the help of False negative (FN), True Negative (TN), True Positive (TP), False Positive (FP). The equations are given below:

$$\text{Accuracy} = \frac{(TP+TN)}{(TP+TN+FP+FN)} \tag{28}$$

$$\text{Sensitivity} = \frac{TP}{(TP+FN)} \tag{29}$$

$$\text{Specificity} = \frac{TN}{(TN+FP)} \tag{30}$$

$$\text{Youden Index} = \text{Sensitivity} + \text{Specificity} - 1 \quad (31)$$

$$\text{Precision} = \frac{TP}{(TP+FP)} \quad (32)$$

$$\text{F1-Score} = \frac{2*Precision*Sensitivity}{Precision+Sensitivity} \quad (33)$$

The results obtained in terms of considered evaluation metrics considered are shown in **Table 2.**

**Table 2.** Classification results of various classifiers and proposed scheme

| Classifier | | Accuracy % | Precision % | Sensitivity % | F1-score% | Youden Index% | Specificity% |
|---|---|---|---|---|---|---|---|
| **Proposed Method (KNN-RF-DT)** | | **97.305** | **97.73** | **97.04** | **97.41** | **94.71** | **97.60** |
| SVM Kernel | RBF | 93.038 | 92.38 | 93.82 | 94.79 | 89.50 | 92.26 |
| | Linear | 85.56 | 85.20 | 86.07 | 85.41 | 70.72 | 85.05 |
| | Polynomial | 89.39 | 88.79 | 90.22 | 90.25 | 80.39 | 88.58 |
| Naïve Bayes | | 81.33 | 81.68 | 80.83 | 81.62 | 63.54 | 81.85 |
| Decision Tree | | 93.157 | 93.58 | 92.80 | 95.45 | 90.98 | 93.51 |
| Neural Network | | 93 | 92.57 | 93.27 | 95.30 | 90.61 | 92.76 |
| KNN | | 94.765 | 94.92 | 94.30 | 94.60 | 89.53 | 95.23 |

From the results shown in Table 2, we can conclude that the proposed method with accuracy 97.305% outperforms the compared classifiers. We used KNN, RF and DT in hybrid ensemble classifier because they give best performance in terms of various evaluation parameters when compared to other classifiers but it increases time complexity due to computation by three individual classifiers. Accuracy of Random Forest is always greater than the Decision tree due to which RF is included in hybrid ensemble classifier. Sensitivity and Specificity parameters are quite comparable percentage-wise. Sensitivity results describe benign tumors, which are good determinative, as it is merely a total positive divide by the total actual benign tumor. Specificity results describe malignant tumors, which are good determinative, as it is merely a total negative divide by total actual malignant tumor. Youden-index describes the maximum difference between true-positive, and false-positive, high Youden-index means true-positive results are quite high as compared to false positive. F1-score describes the balance between precision and Sensitivity, which is vital as there occur uneven class distribution. Precision describes the percentage of actual positive (TP) among the entire predicted positive (TP+FP), which is quite high for the proposed method. Overall, the comparison between the proposed and already existing classification methods, which are implemented in above table is shown below in Fig. 10.

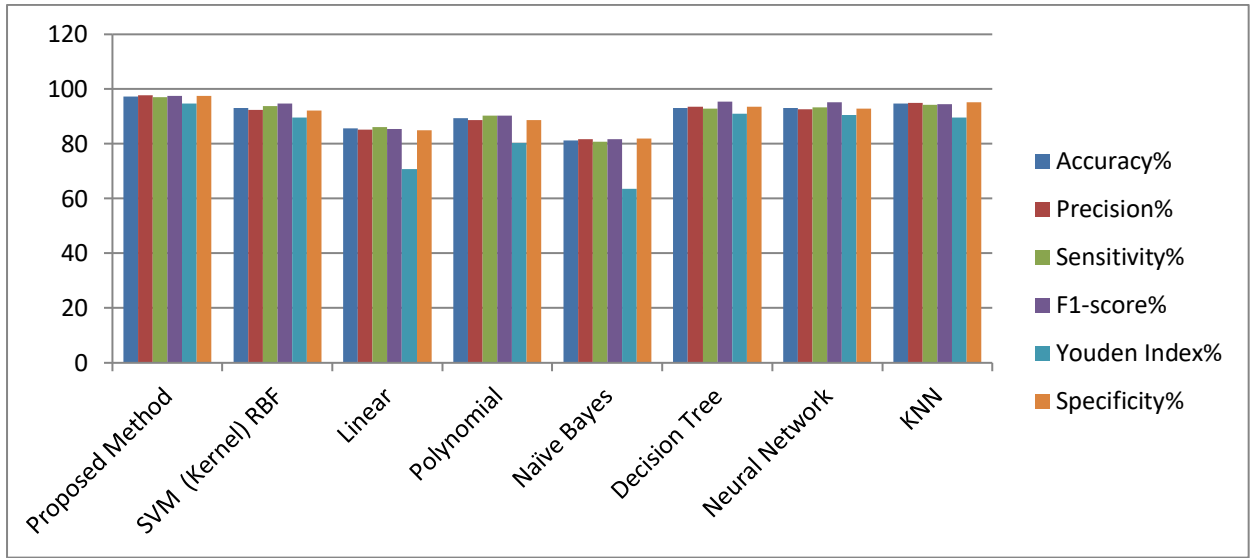

**Fig. 10** Comparison of proposed and existing classification methods based on performance metrics.

c. Confusion Matrix and GUI

|  Predicted/Actual → | Benign | Malignant |
|---|---|---|
| Benign | 1249 | 29 |
| Malignant | 39 | 1239 |

**Fig. 11** Confusion matrix of the proposed method.

Confusion matrix for proposed method KNN-RF-DT based on majority voting is shown in Fig.11. 1249 of 1278 benign tumors are classified as benign, while 29 tumors as malignant. Besides, 1239 of 1278 malignant tumors are classified as malignant, while 39 tumors as benign. Overall, a good accuracy of 97.305 is obtained using the proposed method.

  Overall, the proposed method gives excellent performance on various evaluation parameters as described above in comparison to existing methods. Thus, the proposed method is novel and effective to work for the classification of benign and malignant brain tumors.

  GUI for this proposed method makes it more user-friendly, which is implemented in MATLAB 2017a itself shown below in Fig.12

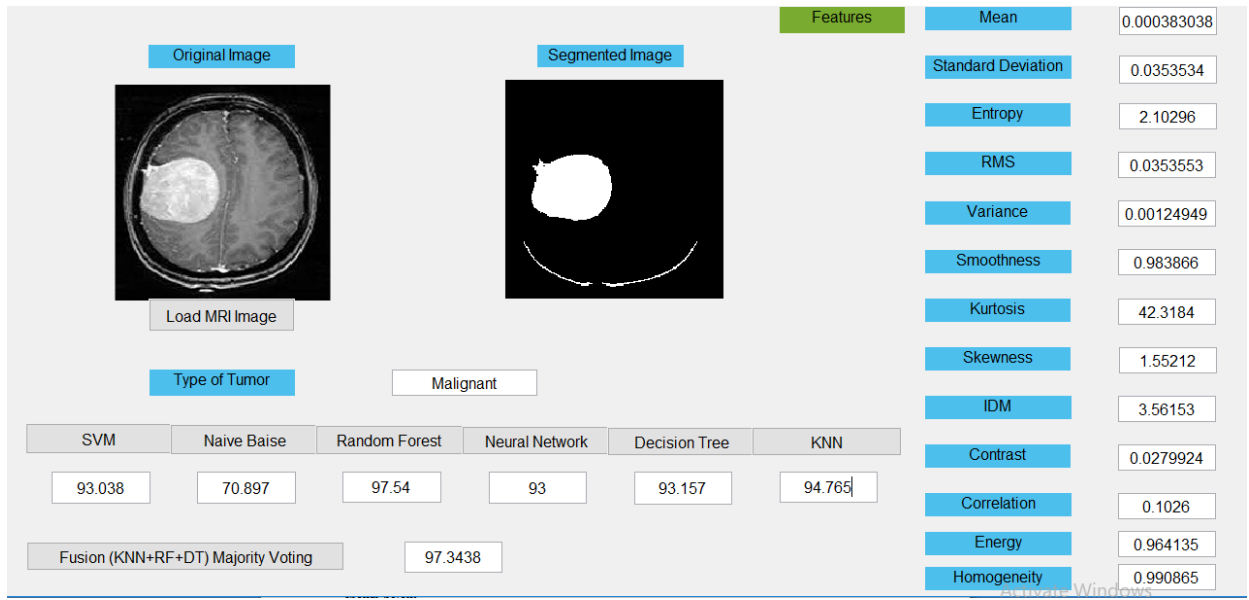

**Fig. 12** GUI for the above-proposed method in Matlab 2017a.

d. Area Calculation for segmented Region

Area calculation for segmented images is shown below in Table 3 using formula:

$$\text{Image, I} = \sum_{w=0}^{200} \sum_{h=0}^{200} [g(0) + g(1)] \tag{34}$$

where Pixels = width (w) * height (h) = 200*200
g(0) = black pixel (digit 0)
g(1) = white pixel (digit1)

$$\text{No. of white pixels, P} = \sum_{w=0}^{200} \sum_{h=0}^{200} [g(1)], \tag{35}$$

where P is number of white pixels and 1Pixel = .264 mm

The formula for area calculation as follow:
$$\text{Area} = [\text{sqrt}(P)*.264] \text{ in mm}^2 \tag{36}$$

**Table 3.** Describes the area calculation of segmented images

| Segmented Images | 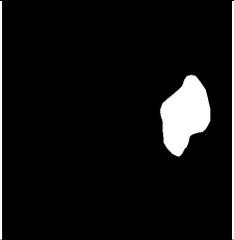 | 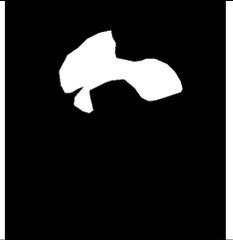 | 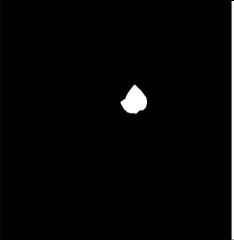 | 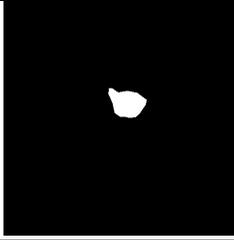 |
|---|---|---|---|---|
| Area (mm$^2$) | 28.6656 | 38.5765 | 12.3235 | 15.2869 |

e. Analysis of Time Complexity

This section presents the time complexity of traditional classifiers [18] and the proposed hybrid ensemble classifier and are shown in Table 4. Further, we are comparing the time complexity of proposed method to that of modern deep learning approaches like convolutional neural network (CNN).

For proposed hybrid ensemble classifier, training time complexity is $O(1+ n^2 p n_{trees} + n^2 p)$ where $n_{trees}$ represents number of trees of random forest, $n$: no. of training samples, $p$: no. of features used. Here we have considered $n_{trees} = 100$; $n = 2172$, which is 85% of total dataset (2556); $p = 13$. Thus, O $(1 + (2172)^2*13*(100+1))$ is equivalent to **O (6.1942e9)**.

Next, we compute training time complexity of deep learning classifiers for instance CNN. CNN consists of an input layer, several convolutional layers, pooling layers, fully connected layers and an output layer. Let us suppose, following represents the architecture of CNN.

Size (pixels *pixels) of the input image being used ($I*I = 200*200$); Size of kernel ($S2*S2 = 7*7$); No. of kernels (N1 =20) in first Convolutional Layer; Pooling Size = 2*2 Pixels with stride = 2 Pixels; Size of kernel ($S3*S3 = 4*4$), No. of kernels ($N2 = 10$) in second Convolutional Layer; Pooling Size = 2*2 Pixels with stride = 2 Pixels.

Here we process input image step by step to extract the number of features that become input to Fully connected layers of CNN. When first convolutional layer kernels are applied we get 200-7+1 by 200-7+1 size with N1=20 matrix i.e. 194*194 size 20 matrices. When we apply pool layer we get 194/2 by 194/2 size with N1=20 matrices i.e. 97*97 size 20 matrices. In second convolution layer, kernels are applied to the output of first convolutional layers and we get 97-4+1 by 97-4+1 size N1*N2 matrices i.e. 94*94 size 200 matrices. When we apply pool layer we get 94/2 by 94/2 size 200 matrices i.e. 47*47 size 200 matrices. Now this 47*47*200 becomes the input to the fully connected layer, which is nothing but the simple Feed-forward back-propagation neural network. So, here training time complexity of the model (Fully connected layers) is $O(nt*(pj + jk))$, where $n=2172$: no. of training set, $t = 1000$: no. of epochs and $p=47*47*200$ (input layer), $j=20$ (hidden layer), $k=2$ (output layer) i.e. O(2172*1000(47*47*200*20+20*2)) = **O(1.9192e13).**

Training time complexity of proposed hybrid model is **O(6.1942e9),** which is quite less as compared to the training time complexity of deep learning classifiers like CNN **O(1.9192e13)** for same set of parameters. In reality, training time complexity of CNN will be much more than what we have calculated here. Because in present calculation, we have taken same number of training dataset for proposed and CNN but in reality number of dataset images would be much higher than 2172 for CNN method and so is the number of neurons required in hidden layer as well.

Overall, we conclude that the proposed hybrid ensemble classifier provides good accuracy at the expense of increased the training time complexity in comparison to traditional classifiers like DT, SVM, and KNN etc. However, it has significantly less training time complexity as compared to modern deep learning methods like CNN and provides the comparative accuracy.

**Table.4.** Describes the time complexity of traditional classifiers

| Complexity | Training | Prediction |
|---|---|---|
| KNN-RF-DT | $O(1+ n^2 p n_{trees} + n^2 p)$ where $n_{trees} = 100$ i.e. number of trees of random forest, $n = 2172$: no. of training samples for model, which is 85% of total dataset (2556), $p = 13$: no. of features used. | $O(np+pn_{trees}+p)$ |
| Decision Tree | $O(n^2 p)$ | $O(p)$ |
| SVM (rbf) | $O(n^2 p+n^3)$ | $O(n_{sv} p)$, where $n_{sv} = 1$ |
| KNN | $O(1)$ | $O(np)$ |
| Neural Network | $O(nt*(pj + jk))$ where $t = 1000$: no. of epochs and $p=13$(input layer), $j=20$(hidden layer), $k=2$ (output layer) | $O(pj+jk)$ where $p = 13$ (input layer), $j = 20$ (hidden layer), $k = 2$ (output layer) |
| Naïve Bayes | $O(np)$ | $O(p)$ |
| Random Forest | $O(n^2 p n_{trees})$ | $O(p n_{trees})$ |

## 5. Conclusion

The proposed work aims at improving the performance of traditional classifiers. As traditional classifiers have an advantage over deep learning algorithms because they require small datasets for training and have low computational time complexity. Image is segmented using otsu's method, features are extracted by using SWT+PCA+GLCM, and finally, classification is done based on hybrid ensemble classifier KNN-RF-DT. The proposed method is novel and useful as it outperforms the already existing methods based on machine learning. Experiments are conducted with software MATLAB 2017a with a personal computer of 4 GB memory, Windows 10 64-bit operating system, and Intel (R) Core (TM) i3-6006U CPU @ 2.00 GHz. Overall, proposed method achieved accuracy of 97.305%, precision 97.73%, specificity 97.60%, Sensitivity 97.04%, Youden-index 94.71%, and F1-score 97.41% which indicates its authenticity over medical images. In future, other hybridization ideas will be investigated like Neural Network-SVM, Neural Network-KNN, Neural Network-RF, Neural Network-DT and Neural Network – Naïve Bayes to further improve the accuracy.